\documentclass{bmvc2k}

\usepackage{tikz}
\usepackage{pgfplots}

\usepackage{amssymb}
\usepackage{amsmath}

\title{Memory-efficient Global Refinement of Decision-Tree Ensembles and its Application to Face Alignment}

\addauthor{Nenad Marku\v{s}}{nenad.markus@fer.hr}{1}
\addauthor{Ivan Gogi\'c}{ivan.gogic@fer.hr}{1}
\addauthor{Igor S. Pand\v{z}i\'c}{igor.pandzic@fer.hr}{1}
\addauthor{J\"orgen Ahlberg}{jorgen.ahlberg@liu.se}{2}

\addinstitution{
 Faculty of Electrical Engineering and Computing\\
 University of Zagreb\\
 Unska 3, 10000 Zagreb, Croatia
}
\addinstitution{
Computer Vision Laboratory\\
Dept. of Electrical Engineering\\
Link\"oping University\\
SE-581 83 Link\"oping, Sweden
}

\runninghead{Marku\v{s}, Gogi\'c, Pand\v{z}i\'c, Ahlberg}{Memory-efficient Global Refinement}

\begin{document}

\maketitle

\begin{abstract}
Ren et al. \cite{lbf} recently introduced a method for aggregating multiple decision trees into a strong predictor by interpreting a path taken by a sample down each tree as a binary vector and performing linear regression on top of these vectors stacked together.
They provided experimental evidence that the method offers advantages over the usual approaches for combining decision trees (random forests and boosting).
The method truly shines when the regression target is a large vector with correlated dimensions, such as a 2D face shape represented with the positions of several facial landmarks.
However, we argue that their basic method is not applicable in many practical scenarios due to large memory requirements.
This paper shows how this issue can be solved through the use of quantization and architectural changes of the predictor that maps decision tree-derived encodings to the desired output.
\end{abstract}

\section{Introduction}
\label{sec:intro}
Decision trees \cite{cart} are a tried-and-true machine learning method with a long tradition.
They are especially powerful when combined in an ensemble \cite{rand_forests}
(the outputs of multiple trees are usually summed together).
On certain problems, these approaches are still competitive \cite{doweneed,bestqm}.

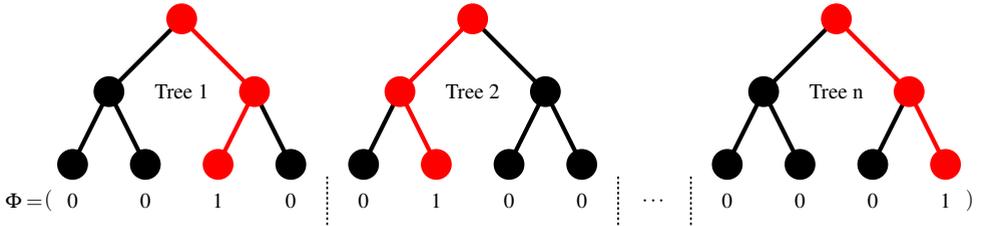
\begin{figure}
	\centering
	\resizebox{1.0\textwidth}{!}
	{
		\begin{tikzpicture}

			\node at (-1, -0.5)
				{\Huge$\Phi=$};
			\node at (0, -0.5)
				{\Huge$($};
			\node at (38, -0.5)
				{\Huge$)$};

			\node at (5.5, 4)
				{\Huge Tree 1};
			\draw[-, line width=5]
				(1, 1) -- (2.5, 4);
			\draw[-, line width=5]
				(4, 1) -- (2.5, 4);
			\draw[-, line width=5, red]
				(7, 1) -- (8.5, 4);
			\draw[-, line width=5]
				(10, 1) -- (8.5, 4);
			\draw[-, line width=5]
				(2.5, 4) -- (5.5, 7);
			\draw[-, line width=5, red]
				(8.5, 4) -- (5.5, 7);
			\draw[-, fill, line width=2]
				(1, 1) ellipse (0.6);
			\draw[-, fill, line width=2]
				(4, 1) ellipse (0.6);
			\draw[-, fill, line width=2, red]
				(7, 1) ellipse (0.6);
			\draw[-, fill, line width=2]
				(10, 1) ellipse (0.6);
			\draw[-, fill, line width=2]
				(2.5, 4) ellipse (0.6);
			\draw[-, fill, line width=2, red]
				(8.5, 4) ellipse (0.6);
			\draw[-, fill, line width=2, red]
				(5.5, 7) ellipse (0.6);
			\node at (1, -0.5)
				{\Huge$0$};
			\node at (4, -0.5)
				{\Huge$0$};
			\node at (7, -0.5)
				{\Huge$1$};
			\node at (10, -0.5)
				{\Huge$0$};

			\draw[dashed, line width=2]
				(11.5, -1.5) -- (11.5, 0.5);

			\node at (17.5, 4)
				{\Huge Tree 2};
			\draw[-, line width=5]
				(13, 1) -- (14.5, 4);
			\draw[-, line width=5, red]
				(16, 1) -- (14.5, 4);
			\draw[-, line width=5]
				(19, 1) -- (20.5, 4);
			\draw[-, line width=5]
				(22, 1) -- (20.5, 4);
			\draw[-, line width=5, red]
				(14.5, 4) -- (17.5, 7);
			\draw[-, line width=5]
				(20.5, 4) -- (17.5, 7);
			\draw[-, fill, line width=2]
				(13, 1) ellipse (0.6);
			\draw[-, fill, line width=2, red]
				(16, 1) ellipse (0.6);
			\draw[-, fill, line width=2]
				(19, 1) ellipse (0.6);
			\draw[-, fill, line width=2]
				(22, 1) ellipse (0.6);
			\draw[-, fill, line width=2, red]
				(14.5, 4) ellipse (0.6);
			\draw[-, fill, line width=2]
				(20.5, 4) ellipse (0.6);
			\draw[-, fill, line width=2, red]
				(17.5, 7) ellipse (0.6);
			\node at (13, -0.5)
				{\Huge$0$};
			\node at (16, -0.5)
				{\Huge$1$};
			\node at (19, -0.5)
				{\Huge$0$};
			\node at (22, -0.5)
				{\Huge$0$};

			\draw[dashed, line width=2]
				(23.5, -1.5) -- (23.5, 0.5);
			\node at (25, -0.5)
				{\Huge$\cdots$};
			\draw[dashed, line width=2]
				(26.5, -1.5) -- (26.5, 0.5);

			\node at (32.5, 4)
				{\Huge Tree n};
			\draw[-, line width=5]
				(28, 1) -- (29.5, 4);
			\draw[-, line width=5]
				(31, 1) -- (29.5, 4);
			\draw[-, line width=5]
				(34, 1) -- (35.5, 4);
			\draw[-, line width=5, red]
				(37, 1) -- (35.5, 4);
			\draw[-, line width=5]
				(29.5, 4) -- (32.5, 7);
			\draw[-, line width=5, red]
				(35.5, 4) -- (32.5, 7);
			\draw[-, fill, line width=2]
				(28, 1) ellipse (0.6);
			\draw[-, fill, line width=2]
				(31, 1) ellipse (0.6);
			\draw[-, fill, line width=2]
				(34, 1) ellipse (0.6);
			\draw[-, fill, line width=2, red]
				(37, 1) ellipse (0.6);
			\draw[-, fill, line width=2]
				(29.5, 4) ellipse (0.6);
			\draw[-, fill, line width=2, red]
				(35.5, 4) ellipse (0.6);
			\draw[-, fill, line width=2, red]
				(32.5, 7) ellipse (0.6);
			\node at (28, -0.5)
				{\Huge$0$};
			\node at (31, -0.5)
				{\Huge$0$};
			\node at (34, -0.5)
				{\Huge$0$};
			\node at (37, -0.5)
				{\Huge$1$};

		\end{tikzpicture}
	}
	\caption
	{
		The process of generating a sparse feature vector $\Phi$ with an ensemble of $n$ decision trees.
		The path that a sample takes through each tree is drawn in red.
		Exactly $n$ components of $\Phi$ are set to $1$ based on these paths.
		The rest are set to $0$.
	}
	\label{fig:treeinds}
\end{figure}

A nice property of decision-tree ensembles is that the method easily deals with multidimensional prediction
(e.g., in multi-class classification).
This is achieved by placing a vector in the leaf node of each tree.
This means that the multidimensional output $\Delta(x)$ for the input sample $x$ is computed as $\Delta(x)=\sum_i\mathbf{w}_i$,
where the $i$th vector is output by the $i$th tree: $\mathbf{w}_i=\text{Tree}_i(x)$.
Ren et al. \cite{lbf,global} interpret this computation as a linear projection step:
\begin{equation}\label{eq:linproj}
\Delta(x)=
\sum_i\text{Tree}_i(x)=
\mathbf{W}\cdot\Phi(x)
,
\end{equation}
where $\mathbf{W}$ is a large matrix that contains as columns the leaf-node vectors of the trees and $\Phi(x)$ is a sparse vector that indicates which columns of $\mathbf{W}$ should be summed together in order to obtain the prediction for the sample $x$.
See Figure \ref{fig:treeinds} for an illustration that shows how to obtain $\Phi(x)$.
This interpretation enabled Ren et al. to learn an efficient method for face alignment by jointly refining the outputs of multiple decision trees \cite{lbf}
(more details later).

If the ensemble has $n$ trees of depth equal to $d$ and the dimension of the output is $o$,
then the matrix $\mathbf{W}$ has $n \cdot 2^d \cdot o$ parameters.
This number can be quite large in a practical setting.
We show how to replace Equation \eqref{eq:linproj} with a more memory-friendly computation:
first, we investigate two different methods with a reduced number of coefficients;
second, we show that the remaining coefficients can be further compressed with quantization.
We apply our ideas to a face-alignment problem and compare with several recently published methods. 

To summarize, the goal of this paper is twofold:
\begin{itemize}
	\item we improve on the previous work of Ren et al. \cite{lbf,lbf-tip} by significantly reducing the memory requirements of their methods with no loss in accuracy;
	\item we describe a new face-alignment method that is reliable, fast and memory-efficient, which makes it especially suitable for use on mobile devices and other hardware with limited resources.
\end{itemize}

\section{Memory-efficient global refinement of tree ensembles}\label{sec:method}
A way to compress $\mathbf{W}\in\mathbb{R}^{o\times(n\cdot 2^d)}$ is to express it as a product of $\mathbf{W}_2\in\mathbb{R}^{o\times r}$ and $\mathbf{W}_1\in\mathbb{R}^{r\times(n\cdot 2^d)}$:
$\mathbf{W}=\mathbf{W}_2\cdot\mathbf{W}_1$.
Of course, $r$ has to be smaller than $o$.
For classification, we can attempt to learn these matrices using gradient descent.
For regression, besides gradient descent, we can also use the reduced-rank regression (RRR) framework \cite{rrr}.

Another possible path to improving memory issues is to replace linear regression for computing $\Delta$ from $\Phi$ (Equation \eqref{eq:linproj}) with a neural network (NN).
One architecture that we found to work well in our experiments is
\begin{equation}\label{eq:nneq}
\Delta(x)=
\mathbf{W}_3\cdot\text{\texttt{tanh}}\left(\mathbf{W}_2\cdot\text{\texttt{tanh}}\left(\mathbf{W}_{1}\cdot\Phi(x)\right)\right)
,
\end{equation}
where $\mathbf{W}_3\in\mathbb{R}^{o\times 2r}$, $\mathbf{W}_2\in\mathbb{R}^{2r\times r}$, $\mathbf{W}_1\in\mathbb{R}^{r\times(n\cdot 2^d)}$ and \texttt{tanh} is the elementwise hyperbolic-tangent nonlinearity.
See Figure \ref{fig:nn} for an illustration. We agree that there might be better design choices. However, we leave these question to practitioners (as the proposed framework is generic) and experimentally test the performance of the architecture specified with Equation \ref{eq:nneq}.
\input{nn.fig}
The matrices $\mathbf{W}_1$, $\mathbf{W}_2$ and $\mathbf{W}_3$ can be learned with gradient descent through the use of backpropagation.
The presented NN architecture improves memory issues only if $r$ can be made reasonably small.
Also, note that modern tricks that we do not use here (e.g., highway connections \cite{highway} and batch normalization \cite{bnorm}) could further boost performance.

Instead of a standard tree-based prediction, which is additive in nature (Equation \eqref{eq:linproj}),
we now have a two-step procedure:
\begin{itemize}
	\item
	produce a compact encoding $\mathbf{e}(x)$, $\mathbf{e}(x)=\sum_i\text{Tree}_i(x)$;
	\item
	decode $\Delta(x)$ from $\mathbf{e}(x)$
	(this can be a nonlinear, non-additive computation, such as a neural network).
\end{itemize}

Our claim is that $r$ can be significantly smaller than $o$ in some interesting problems and that, consequently, this leads to large reduction in storage-related issues for computing $\Delta$ from $\Phi$.
We later back these claims by experimental evidence.

\subsection{Related work}
For the history and applications of reduced-rank regression please see the book by Reinsel and Velu \cite{rrr}.
As this is a well-known and widely-used approach in machine learning and statistics, we will not repeat the details here.

Our research is related to a large body of work in neural-network compression and speed improvement (e.g., \cite{speechrecog,speechrecog2,googlenet,exploiting,deepcompr}).
Specifically, the work of Sainath et al. \cite{speechrecog} and other similar approaches suggest to reduce the memory-related issues of machine learning models by introducing a low-dimensional "bottleneck" before the output, as we do in this paper.
However, to the best of our knowledge, we are the first to investigate the use of these techniques in the context of decision-tree ensembles.
We can say that our work is a blend of the interpretation introduced by Ren et al. \cite{lbf} and recent network compression techniques.

In the next section, we apply the proposed techniques to the problem of face alignment. The literature in this area is large and we give more information only on most related methods \cite{microsoft_face_align,lbf-tip,lbf,SDM} there as our work builds on the same core ideas.



\section{Application to face alignment}
Face alignment is the process of finding the shape of the face given its location and size in the image.
Today's most popular methods (e.g., \cite{microsoft_face_align,SDM,lbf}) achieve this through $t=1, 2, \ldots, T$ iterations:
$\mathbf{S}_t=\mathbf{S}_{t-1} + \Delta_t$, where $\Delta_t=\Delta(\mathbf{S}_{t-1}, F)$ is the shape update computed from the previous shape $\mathbf{S}_{t-1}$ and image $F$.
The shape update is usually computed in two steps \cite{SDM,lbf}.
The first step is a feature extraction process $\Phi(\mathbf{S}_{t-1}, F)$ that encodes the image appearance around the shape $\mathbf{S}_{t-1}$.
The second step computes the shape increment via a linear projection: $\Delta_t=\mathbf{W}_t\cdot\Phi(\mathbf{S}_{t-1}, F)$.

For the supervised descent method of Xiong and De la Torre \cite{SDM}, the features $\Phi$ are stacked SIFT descriptors extracted around landmark points $\mathbf{S}_{t-1}$.
The matrices $\{\mathbf{W}_t\}$ are learned from a training set by ordinary least squares regression.
A more complicated method of Ren et al. \cite{lbf} extracts a sparse feature vector around each landmark point from $\mathbf{S}_{t-1}$ with a decision tree ensemble
and concatenates them to obtain $\Phi$.
The binary tests in internal tree nodes are thresholded pixel intensity differences, $F_{x_1,y_1}-F_{x_2,y_2}$.
The simplicity of these tests makes the method very fast in runtime.
The sparse feature vector $\Phi$ computed in this manner is very large 
(tens of thousands of dimensions in practical settings).
Consequently, the matrices $\{\mathbf{W}_t\}$ have a lot of parameters (weights).
Our goal is to experimentally show that we can replace linear projection $\Delta_t=\mathbf{W}_t\cdot\Phi_t$ with a more memory-friendly computation. For a more in-depth review on recent advances in face alignment, we refer the reader to~\cite{jin2017face}.

We base all our experiments on the 300 Faces in-the-Wild framework (300W for short, \cite{300w}),
which is a standard procedure for benchmarking face-alignment systems.
Each face is represented with a $136$-dimensional vector ($68$ landmark points, two coordinates each).
The dataset was constructed to examine the ability of face-alignment systems to handle naturalistic, unconstrained face images.
It covers many different variations of unseen subjects, pose, expression, illumination, background, occlusion and image quality.
These features have made the 300W dataset a \textit{de facto} standard for benchmarking face-alignment systems.
The dataset is partitioned into the training and testing subsets according to a predefined protocol.
We follow these guidelines closely.

\subsection{Learning setup}
To augment the training data, we apply the following transformations $20$ times for each face:
flip the image with probability equal to $0.5$ (faces are symmetric),
rotate the face for a random angle between $-20$ and $20$ degrees,
randomly perturb the scale and the position of the face bounding box.
Overall, this results in a training set of around $50\;000$ samples.

The number of trees per landmark point is set to $5$ and the depth of each individual tree is $5$.
It can be seen from the setup that the dimensionality of vector $\Phi$ is $68\cdot 5\cdot 2^5=10\;880$.
Initially, each tree is learned to predict the position of its landmark point from a local patch around it.
Next, the values in the leaf nodes are discarded and we learn the function $\Delta(x)$ using the method proposed by Ren et al. \cite{lbf} (Equation \eqref{eq:linproj}) or one of the methods described in Section \ref{sec:method}.
The total number of stages is also set to $5$.
These hyper-parameters follow the ones proposed in the original LBF (Local Binary Features) paper \cite{lbf}.
The initial shape is set to be the mean face computed on the training data.

For reduced-rank regression and the neural-net approach, our preliminary experiments showed that it is beneficial to increase the size of the bottleneck as the shape estimation progresses through the stages.
The following values lead to good results: $r=16, 24, 32, 40, 48$.
This is based on the heuristic that the transformations/deformations in the first iterations are more rigid and, thus, require less parameters to be modeled reliably.
On the other hand, the later iterations require a considerable number of parameters to fine-tune the landmark positions.
These values were tuned on the training/validation set partitions (i.e., the testing set was left out of this experimentation).
However, the search was not exhaustive and there is a possibility that some other values might lead to better results.

The projection matrix for each stage of the original LBF method is learned with LIBLINEAR \cite{liblinear}
(we closely follow their recommendations \cite{lbf}).
To get the optimal reduced-rank regression matrices, we use BLAS and LAPACK since the process involves eigenvalue problems, matrix multiplications and inversions \cite{rrr}.
For the backpropagation-based methods, we use the Torch7 framework as it implements a very efficient \texttt{nn.SparseLinear()} module for computing products of sparse vectors and matrices.
The SGD learning rate is initially set to $\eta=1$ and halved each time the validation loss saturates.
The minibatch size is $128$.
The learning for all methods is quite fast in the described setting (a few hours).

To make the exposition more compact and reduce the clutter in graphs/plots, we introduce the following abbreviations:
LL for the basic LBF method \cite{lbf}, RRR for reduced-rank regression \cite{rrr}, RRR-BP for reduced-rank regression learned with backpropagation and NN for the neural networks.

\subsection{First results and the effects of random initializations}
The average point-to-point Euclidean error for $L=68$ landmark points normalized by the inter-ocular distance
(IOD, measured as the Euclidean distance between the outer corners of the eyes) is used to compare different methods:
\begin{equation}\label{eq:err}
\text{err}=
100\times\frac{1}{L}\sum_{i=1}^L\frac{\sqrt{(x_i-x_i^*)^2 + (y_i-y_i^*)^2}}{\text{IOD}}
\;\;[\%]
,
\end{equation}
where $(x_i, y_i)$ and $(x_i^*, y_i^*)$ are the estimated and ground-truth locations of the $i$th landmark.

It is known that averaging the results of several initializations (slight perturbations in position and scale of the face bounding box) can improve face alignment \cite{microsoft_face_align}.
We did not notice high gains in alignment accuracy (Equation \eqref{eq:err}).
However, the method reduces the jitter when processing videos or a webcam stream.
This significantly affects the subjective quality of the face-alignment system.
Thus, we also report the results for different numbers of random initializations, $p$.

Table \ref{tbl:avgerr} contains normalized errors averaged over the common and the challenging testing subsets of the 300W dataset
(as the name implies, these subsets differ in difficulty).
\begin{table}
	\centering
		\begin{tabular}{| c || c | c | c |}
			\hline
			$p$	&	1	&	7	&	15	\\
			\hline
			\hline
			LL	&	4.39	&	4.19	&	4.19	\\
			\hline
			NN	&	4.39	&	3.96	&	3.96	\\
			\hline
			RRR	&	4.39	&	4.03	&	3.95	\\
			\hline
			RRR-BP	&	4.49	&	4.12	&	4.11	\\
			\hline
		\end{tabular}
		\hspace{2pt}
		\begin{tabular}{| c || c | c | c |}
			\hline
			$p$	&	1	&	7	&	15	\\
			\hline
			\hline
			LL	&	11.69	&	11.42	&	11.19	\\
			\hline
			NN	&	11.59	&	10.59	&	10.19	\\
			\hline
			RRR	&	11.18	&	10.21	&	10.06	\\
			\hline
			RRR-BP	&	12.72	&	12.02	&	11.59	\\
			\hline
		\end{tabular}
	\caption
	{
		Average errors for the common (left) and the challenging (right) subsets of the 300W dataset for different number of random initializations, $p$.
	}
	\label{tbl:avgerr}
\end{table}
It can be seen that random perturbations slightly increase the accuracy for all methods.
All competing methods perform approximately the same, i.e., achieve similar average errors.

These results lead to the conclusion that NN- and RRR-based methods are viable alternatives to the basic method of Ren et al. \cite{lbf}.
Moreover, these methods reduce storage/bandwidth requirements by approximately three times
(from $27$ to $9$ megabytes).
However, this reduction may not be sufficient for some applications (such as mobile-phone and web programs with bandwidth constraints).
Thus, the possibility of quantizing the $\mathbf{W}_1$ matrix is explored next.

\subsection{Effects of coefficient quantization}
Quantization is a straightforward way to improve storage requirements.
In all the discussed methods, the matrices that multiply the (very large and sparse) input vector $\Phi$ contain the most parameters.
Thus, these are most interesting to compress with quantization.
In the case of face alignment, these matrices are "long and flat", i.e., they have much more columns than rows.
We found empirically that the following scheme gives good results:
each row of the matrix is quantized to $q$ bits per weight/coefficient with a nonlinear quantizer.
The centroids are computed with the standard $k$-means clustering algorithm where $k=2^q$.
Separate quantizers are used for each row to increase the expressive power without requiring a too-large quantization table.

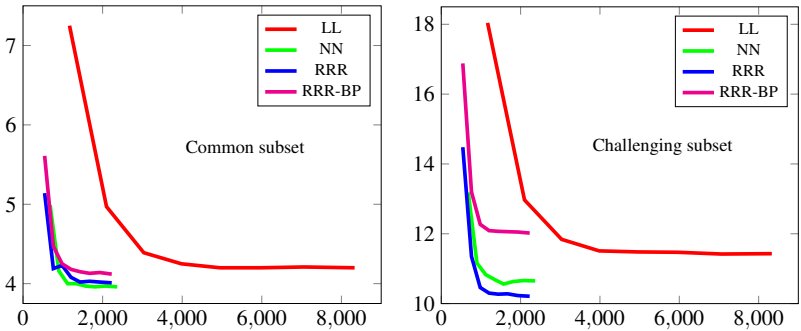
\begin{figure}[t]
	\centering
	\pgfplotsset{every tick label/.append style={font=\tiny}}
\resizebox{0.4\textwidth}{!}
{
	\begin{tikzpicture}
		\node at (4.25, 3)
			{Common subset};
		\begin{axis} [
			ticklabel style = {font=\large},
			xmin=0, xmax=9000,
			ymin=3.75, ymax=7.5,
			legend pos=north east
		]
			\addplot[color=red, line width=2]
				coordinates {
					(8332, 4.20)(7059, 4.21)(5960, 4.20)(4949, 4.20)(3980, 4.25)(3034, 4.39)(2098, 4.97)(1168, 7.25)
				};
			\addplot[color=green, line width=2]
				coordinates {
					(2367, 3.96)(2067, 3.97)(1808, 3.96)(1570, 3.97)(1343, 4.00)(1120, 4.00)(900, 4.16)(681, 4.99)
				};
			\addplot[color=blue, line width=2]
				coordinates {
					(2229, 4.01)(1930, 4.02)(1671, 4.03)(1433, 4.02)(1205, 4.08)(982, 4.23)(762, 4.19)(543, 5.14)
				};
			\addplot[color=magenta, line width=2]
				coordinates {
					(2229, 4.12)(1930, 4.14)(1671, 4.13)(1433, 4.15)(1205, 4.18)(982, 4.25)(762, 4.47)(543, 5.61)
				};
			\legend{\small LL,\small NN,\small RRR,\small RRR-BP}
		\end{axis}
	\end{tikzpicture}
}
\hspace{1pt}
\resizebox{0.41\textwidth}{!}
{
	\begin{tikzpicture}
		\node at (4.25, 3)
			{Challenging subset};
		\begin{axis} [
			ticklabel style = {font=\large},
			xmin=0, xmax=9000,
			ymin=10, ymax=18.5,
			legend pos=north east
		]
			\addplot[color=red, line width=2]
				coordinates {
					(8332, 11.43)(7059, 11.42)(5960, 11.47)(4949, 11.48)(3980, 11.51)(3034, 11.84)(2098, 12.97)(1168, 18.04)
				};
			\addplot[color=green, line width=2]
				coordinates {
					(2367, 10.65)(2067, 10.66)(1808, 10.63)(1570, 10.56)(1343, 10.69)(1120, 10.83)(900, 11.16)(681, 13.18)
				};
			\addplot[color=blue, line width=2]
				coordinates {
					(2229, 10.21)(1930, 10.23)(1671, 10.28)(1433, 10.27)(1205, 10.30)(982, 10.46)(762, 11.35)(543, 14.48)
				};
			\addplot[color=magenta, line width=2]
				coordinates {
					(2229, 12.02)(1930, 12.05)(1671, 12.06)(1433, 12.07)(1205, 12.09)(982, 12.27)(762, 13.20)(543, 16.88)
				};
			\legend{\small LL,\small NN,\small RRR,\small RRR-BP}
		\end{axis}
	\end{tikzpicture}
}
	\caption
	{
		Storage-vs-error tradeoff for the common and challenging subsets of 300W:
		abscissae are in kilobytes and ordinates are normalized errors.
		The results on the "common" subset are in the left graph and the results on the "challenging" subset are in the right one.
	}
	\label{fig:mem-vs-err}
\end{figure}

Parameter quantization leads to significant memory savings.
The exact numerical values can be seen in Table \ref{tbl:qmem} (the RRR-BP method is left out since it has the same memory usage as the ordinary RRR).
\begin{table}
	\centering
		\begin{tabular}{| c || c | c | c | c | c |}
			\hline
			$q$	&	no quantization	&	8	&	6	&	4	&	2	\\
			\hline
			\hline
			LL	&	27901	&	8332	&	5960	&	3980	&	2098	\\
			\hline
			NN	&	6927	&	2367	&	1808	&	1343	&	900	\\
			\hline
			RRR	&	6790	&	2229	&	1671	&	1205	&	762	\\
			\hline
		\end{tabular}
	\caption
	{
		Memory requirements for the packed and quantized regression structures.
		The value $q$ shows the number of bits per weight and the rest of table entries are in kilobytes.
	}
	\label{tbl:qmem}
\end{table}
However, it is not clear how much accuracy degradation is introduced with quantization.
Figure \ref{fig:mem-vs-err} provides more insight into this issue.
The effects are similar for all values of applied randomly perturbed initializations, $p$.
We can see that all the architectures introduced in this paper outperform the basic LBF method (LL, \cite{lbf}) when quantization is used.
We hypothesize that this is due to the fact that only the projection matrices (the $\mathbf{W}_1$ matrices; see Section \ref{sec:method}) are quantized in the implemented scheme.
In other words, parts of the NN- and RRR-based frameworks remain unquantized in all scenarios and it seems that this leads to the preservation of important information.

Overall, quantized RRR- and NN-based methods with $q=4$ and $p=7$ significantly improve over the basic LBF-based shape regression approach introduced by Ren et al. \cite{lbf}.
The storage requirements are reduced by approximately $21$ times (from $27$MB to $1.3$MB) without loss of accuracy.
Note that the gain would be even more significant in the case of a larger face model (more trees, more facial landmarks).

\subsection{Processing speed analysis}
Processing speed results can be seen in Table \ref{tbl:falignspeed}.
\begin{table}
	\centering
	\begin{tabular}{| c || c | c | c | c |}
		\hline
		$p$	&	1	&	3	&	7	&	15	\\
		\hline
		\hline
		LL	&	0.26	&	0.79	&	1.88	&	4.01	\\
		\hline
		NN	&	0.26	&	0.75	&	1.73	&	3.57	\\
		\hline
		RRR	&	0.22	&	0.60	&	1.43	&	2.99	\\
		\hline
	\end{tabular}
	\caption
	{
		Average times in milliseconds required to process one face region for different number of random initializations, $p$.
		CPU: 3.4GHz Core i7-2600.
	}
	\label{tbl:falignspeed}
\end{table}
We can see that all tested methods are extremely fast.
For example, with $p=1$ over $3000$ FPS processing speeds can be achieved.

The measurements also show a slight advantage of RRR-based method over the other two.
This can be attributed to fewer floating point operations when transforming the sparse feature vector to the face shape.
However, this advantage can be beneficial only to a certain point since all methods use the same mechanism to produce the sparse feature vector through pixel-intensity comparison binary tests, which amounts for the majority of computations.

\subsection{Comparison to the other methods}
Table \ref{tbl:300w-cmp} compares the RRR ($q=4$) with recent methods
\cite{zhuetal,DRMF,microsoft_face_align,RCPR,SDM,smithetal,zhaoetal,GN-DPM,CFAN,ERT,lbf,TDCNN,cGPRT,CFSS,DCR}.
The processing speeds are based on the results from \cite{lbf-tip}. The unknown/missing values are denoted with a dash: "--".
Other configurations discussed earlier in the experimental sections, such as those produced with NN- and LL-based approaches and different quantization levels are not included for clarity of presentation due to the fact that all achieve similar error rates.
\begin{table}
	\centering
	\resizebox{0.8\textwidth}{!}
	{
		\begin{tabular}{| c || c | c | c || c |}
			\hline
			Method	&	Common	&	Challenging	&	Full set	&	Proc. speed [FPS]	\\
			\hline
			\hline
			Zhu et al. \cite{zhuetal}	&	8.22	&	18.33	&	10.20	&	--	\\
			\hline
			DRMF \cite{DRMF}	&	6.65	&	19.79	&	9.22	&	--	\\
			\hline
			ESR \cite{microsoft_face_align}	&	5.28	&	17.00	&	7.58	&	120	\\
			\hline
			RCPR \cite{RCPR}	&	6.18	&	17.26	&	8.35	&	80	\\
			\hline
			SDM \cite{SDM}	&	5.57	&	15.40	&	7.50	&	70	\\
			\hline
			Smith et al. \cite{smithetal}	&	--	&	13.30	&	--	&	--	\\
			\hline
			Zhao et al. \cite{zhaoetal}	&	--	&	--	&	6.31	&	--	\\
			\hline
			GN-DPM \cite{GN-DPM}	&	5.78	&	--	&	--	&	--	\\
			\hline
			CFAN \cite{CFAN}	&	5.50	&	--	&	--	&	44	\\
			\hline
			ERT \cite{ERT}	&	--	&	--	&	6.40	&	1000	\\
			\hline
			LBF \cite{lbf}	&	4.95	&	11.98	&	6.32	&	320	\\
			\hline
			LBF (fast) \cite{lbf}	&	5.38	&	15.50	&	7.37	&	3100	\\
			\hline
			TDCNN \cite{TDCNN}	&	4.80	&	8.60	&	5.54	&	56	\\
			\hline
			cGPRT \cite{cGPRT}	&	--	&	--	&	5.71	&	--	\\
			\hline
			CFSS \cite{CFSS}	&	4.73	&	9.98	&	5.76	&	25	\\
			\hline
			CFSS (practical) \cite{CFSS}	&	4.79	&	10.92	&	5.99	&	50	\\
			\hline
			DCR \cite{DCR}	&	4.19	&	8.42	&	5.02	&	--	\\
			\hline
			DAN \cite{kowalski2017deep} & 3.19 & 5.24 & 3.59 & 5 \\
			\hline
			\hline
			RRR ($p=1$)	&	4.52	&	11.34	&	5.86	&	3500+	\\
			\hline
			RRR ($p=7$)	&	4.08	&	10.30	&	5.26	&	500+	\\
			\hline
		\end{tabular}
	}
	\caption
	{
		Comparing other approaches to RRR with $q=4$.
		Normalized errors are in the middle three columns.
	}
	\label{tbl:300w-cmp}
\end{table}
It can be seen that the RRR method achieves competitive results, even when compared to recent deep-learning approaches \cite{DCR}. However, it is important to note that different error normalizations are used by researchers making the exact comparison difficult. We used the inter-ocular normalization (distance between the outer corner of the eyes) since there are no pupil annotations on the 300-W data set while some researchers estimated the pupil position for the inter-pupil normalization.

The conclusions from \cite{lbf-tip} apply.
Overall, the regression-based approaches are better than the ASM-based methods.
Methods based on pixel-intensity comparisons (\cite{microsoft_face_align,ERT,lbf}, RRR) and SIFT-based methods \cite{SDM,CFSS} achieve very good results.
The methods based on convolutional neural nets \cite{CFAN,TDCNN,DCR,kowalski2017deep} achieve the best results.
However, the CNN-based methods do not fare so well in terms of processing speed. Specifically, the  DAN~\cite{kowalski2017deep} method achieves the state-of-the-art accuracy, however it can achieve only 5 FPS while running on a CPU (speed measured using the open source implementation from the authors). In order to achieve real-time processing speeds, CNN-based methods depend on the availability of GPUs. 

The proposed RRR- and NN-based methods provide an excellent accuracy-to-speed trade-off.
Also, their memory requirements are quite modest: less than 2MB of data files.
These characteristics make them an excellent candidate for face alignment on mobile devices and embedded hardware.

\begin{figure}
	\center
	\resizebox{1.0\textwidth}{!}
	{
		\includegraphics[height=3cm]{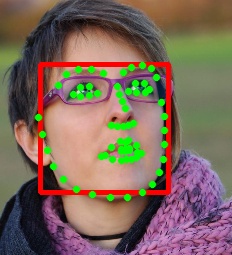}
		\includegraphics[height=3cm]{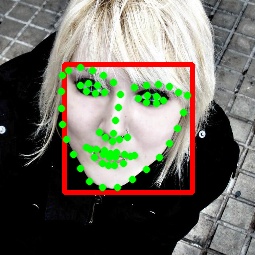}
		\includegraphics[height=3cm]{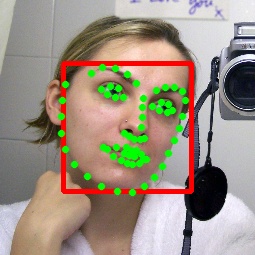}
		\includegraphics[height=3cm]{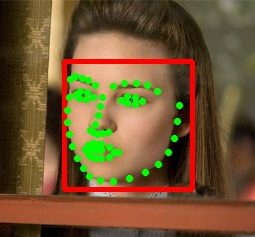}
	}
	\resizebox{1.0\textwidth}{!}
	{
		\includegraphics[height=3cm]{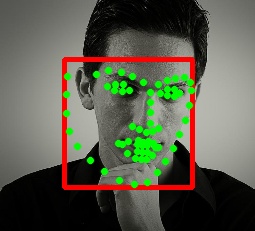}
		\includegraphics[height=3cm]{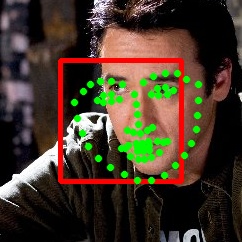}
		\includegraphics[height=3cm]{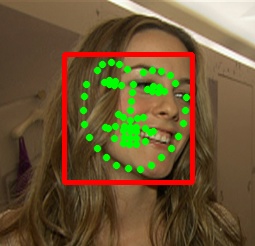}
		\includegraphics[height=3cm]{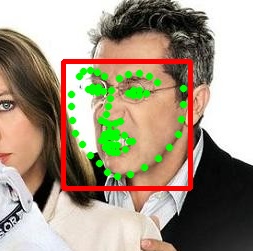}
	}
	\caption
	{
		Some success (top row) and failure (bottom row) cases.
	}
	\label{fig:succfail}
\end{figure}

Some typical success and failure cases for RRR ($q=4$ and $p=7$) can be seen in Figure \ref{fig:succfail}.
The method sometimes fails for points that lie on the face contour.
These images demonstrate this issue in a qualitative way.

\section{Conclusion}
We investigated the possibility of reducing the storage requirements of tree ensembles that predict high-dimensional targets.
The obtained results on the face-alignment task are encouraging.
We recommend to use the backpropagation-based learning of neural architectures that transform the large and sparse tree-derived encodings of the input into the desired output.
This is due to the scalability of the method and its simplicity.
Also, the approach can naturally handle both classification and regression tasks, as well as multi-objective problems.

\bibliography{references}
\end{document}